\documentclass[10pt,twocolumn,letterpaper]{article}

\usepackage[pagenumbers]{cvpr} 

\usepackage{graphicx}
\usepackage{amsmath}
\usepackage{amssymb}
\usepackage{booktabs}
\usepackage{mathtools}
\usepackage[numbers,sort,compress]{natbib}

\usepackage[symbol*]{footmisc}

\usepackage[export]{adjustbox}
\usepackage{multirow}
\usepackage[ruled,vlined]{algorithm2e}
\usepackage[inline]{enumitem}
\usepackage{nicefrac}

\usepackage{ifthen}
\usepackage[normalem]{ulem}
\usepackage{xcolor}

\usepackage{xspace}
\newcommand{\hpvaegan}{HP-VAE-GAN\xspace}

\usepackage[pagebackref,breaklinks,colorlinks]{hyperref}

\usepackage[capitalize]{cleveref}
\crefname{section}{Sec.}{Secs.}
\Crefname{section}{Section}{Sections}
\Crefname{table}{Table}{Tables}
\crefname{table}{Tab.}{Tabs.}

\def\cvprPaperID{***} 
\def\confName{CVPR}
\def\confYear{2022}

\makeatletter
\apptocmd\@maketitle{\teaserfigure{}}{}{}
\makeatother

\begin{document}

\title{\vspace*{-1cm}Diverse Generation from a Single Video Made Possible}

\author{Niv Haim\thanks{These authors contributed equally to this work.} \quad\quad Ben Feinstein\footnotemark[1] \quad\quad Niv Granot \quad\quad Assaf Shocher \\Shai Bagon \quad\quad Tali Dekel \quad\quad Michal Irani \\\\Weizmann Institute of Science, Rehovot, Israel}


\newcommand\teaserfigure{%
\begin{center}
\vspace*{-0.2cm}
\includegraphics[width=\textwidth]{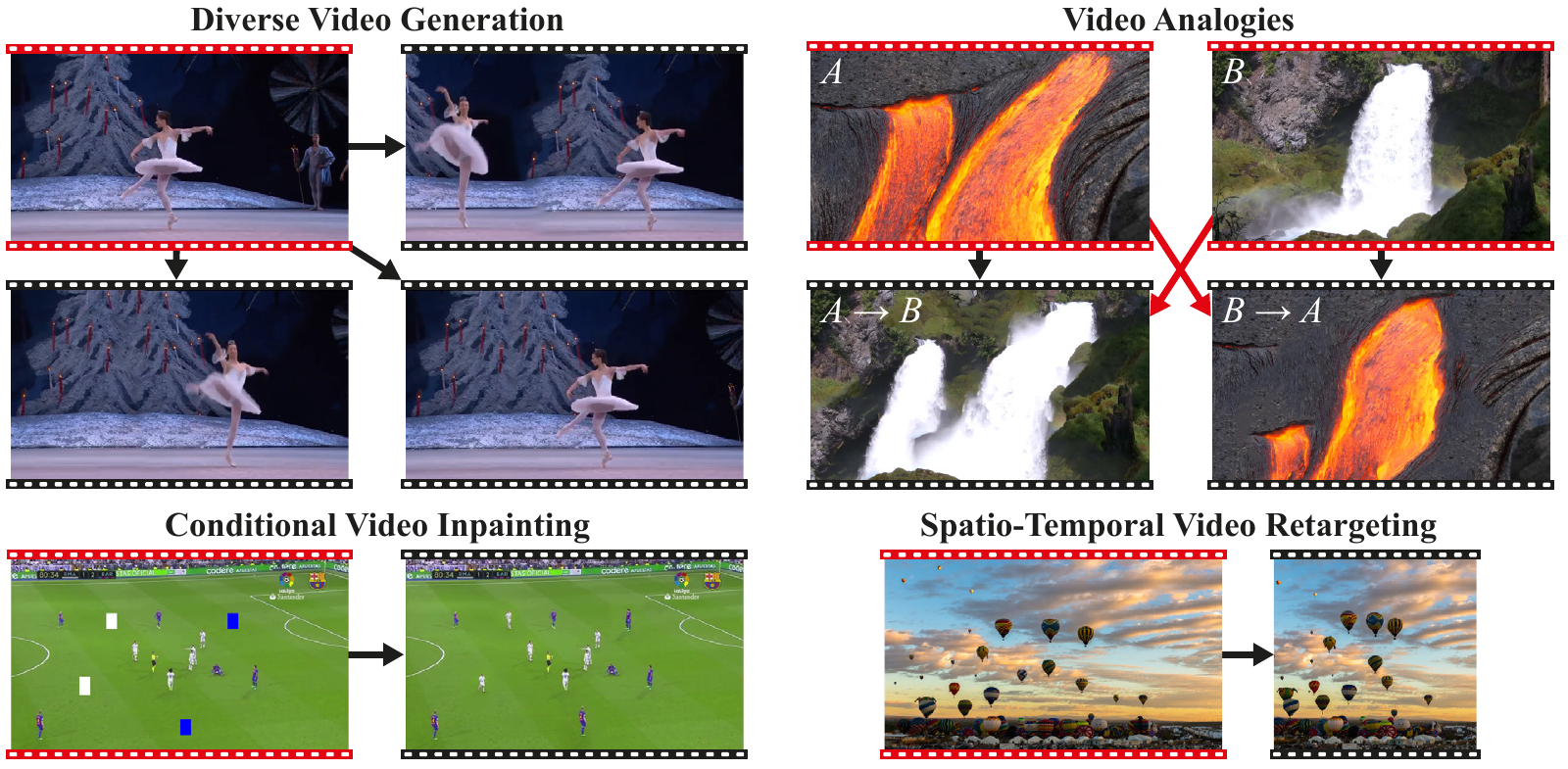} 
\captionof{figure}{We adapt classical patch-based approaches as a better, much faster non-parametric alternative to single video GANs,  for a variety of video generation and manipulation tasks.  As we present video results, the reader is encouraged to start from the project page.
Figures only present single frame examples.}
\vspace*{0.2cm}
\label{fig:teaser}
\end{center}
}

\maketitle

\begin{abstract}
GANs are able to perform generation and manipulation tasks, trained on a single video. However, these single video GANs require unreasonable amount of time to train on a single video, rendering them almost impractical. 
In this paper we question the necessity of a GAN for generation from a single video, and introduce a non-parametric baseline for a variety of generation and manipulation tasks.
We revive classical space-time patches-nearest-neighbors approaches and adapt them to a scalable unconditional generative model, without any learning.
This simple baseline surprisingly outperforms single-video GANs in visual quality and realism (confirmed by quantitative and qualitative evaluations), and is disproportionately faster (runtime reduced from several days to seconds). 
Other than diverse video generation, we demonstrate other applications using the same framework, including video analogies and spatio-temporal retargeting. Our proposed approach is easily scaled to Full-HD videos.
These observations show that the classical approaches, if adapted correctly, significantly outperform heavy deep learning machinery for these tasks. 
This sets a new baseline for single-video generation and manipulation tasks, and no less important --  makes diverse generation from a single video practically possible for the first time. 

\noindent
Project page: \url{https://nivha.github.io/vgpnn}
\end{abstract}

\section{Introduction}
\label{sec:intro}

\begin{figure*}
  \includegraphics[width=\textwidth]{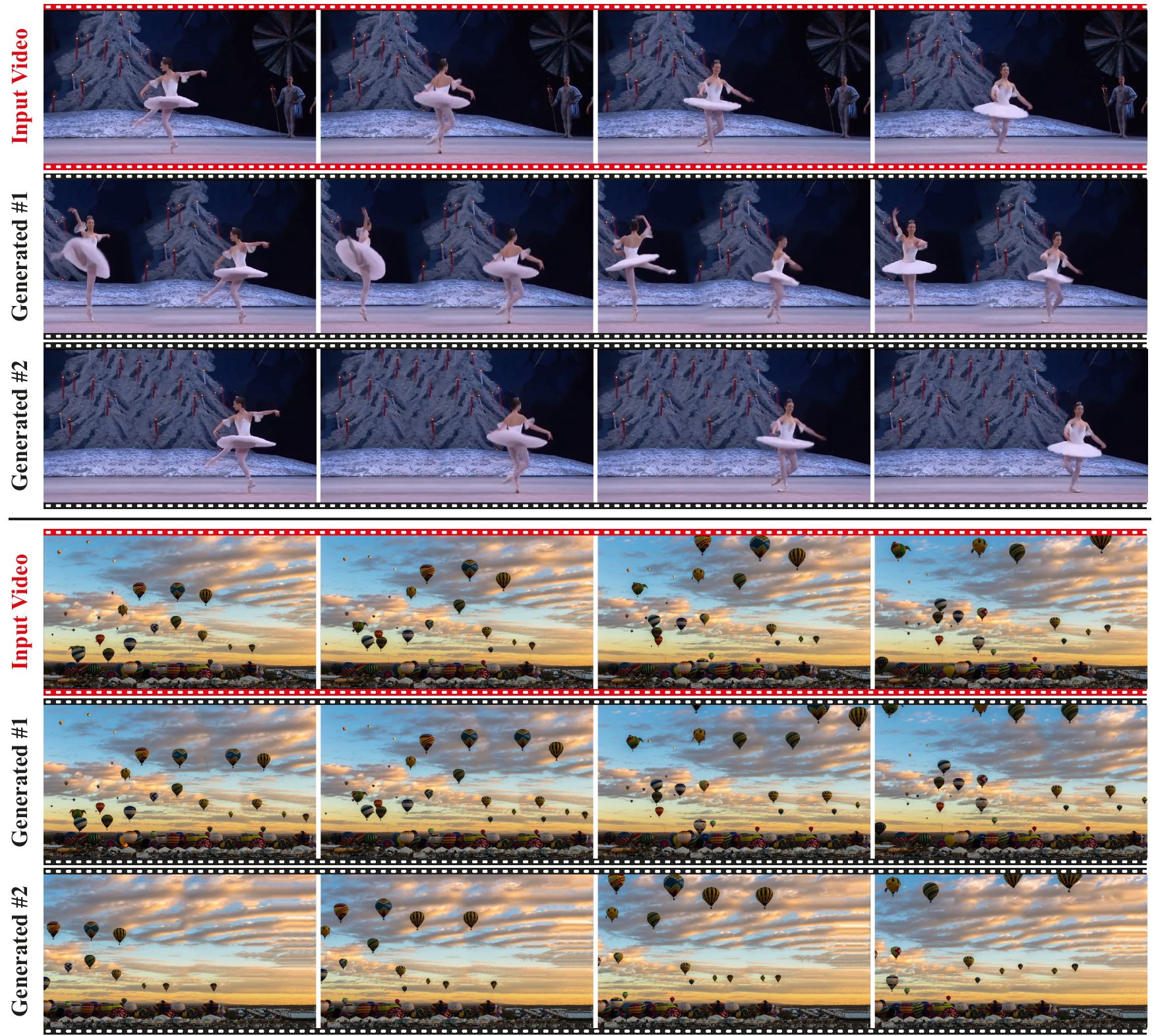}
  \caption{\textbf{Diverse Single Video Generation:} Given an input video (red), VGPNN is able to generate similarly looking videos (black) capturing both appearance of objects as well as their dynamics. Note the high quality of our generated videos. \textit{Please watch the full resolution videos in the project page}.
  }
  \label{fig:generation_examples}
\end{figure*}

Generation and editing of natural videos remain challenging, mainly due to their large dimensionality and the enormous space of motion they span. 
Most modern frameworks train generative models on a large collection of videos, producing high quality results for only a limited class of videos. These include extensions of GANs~\cite{goodfellow2014generative} to video data~\cite{vondrick2016generating, tulyakov2018mocogan, aigner2018futuregan, wang2021inmodegan, saito2017temporal,lee2018stochastic} and video-to-video translation~\cite{wang2018video, wang2019few, mallya2020world, wang2020imaginator, chan2019everybody, yang2020transmomo,bansal2018recycle}, autoregressive sequence prediction~\cite{ballas2015delving,villegas2017decomposing, babaeizadeh2017stochastic, denton2018stochastic, villegas2018hierarchical, villegas2019high, aksan2019stcn, franceschi2020stochastic} and more. 
While externally-trained generative models produce impressive results, they are restricted to the types of video dynamics in their training set. 
On the other side of the spectrum are \emph{single-video GANs}.
These video generative models  train on a \emph{single} input video,
learn its distribution of space-time patches, and are then able to generate a diversity of new videos with the same patch distribution~\cite{gur2020hierarchical, arora2021singan}. 
However, these take very long time to train for each input video, making them applicable to only small spatial resolutions and to very short videos (typically, very few small frames). Furthermore, their output oftentimes shows poor visual quality and noticeable visual artifacts. These shortcomings render existing single-video GANs impractical and unscalable.

Video synthesis and manipulation of a single video sequence based on its distribution of space-time patches dates back to classical  pre-deep learning methods. 
These classical methods demonstrated impressive results for various applications, such as video retargeting~\cite{simakov2008summarizing,rubinstein2008improved,wolf2007videoretarget,krahenbuhl2009disney}, video completion~\cite{wexler2004space,liu2009video,huang2016temporally}, video texture synthesis~\cite{kwatra2003graphcut,kwatra2005texture,kwatra2007texturing,bhat2004flow,fivser2016stylit, jamrivska2015lazyfluids} and more.
With the rise of deep-learning, these methods gradually, perhaps unjustifiably, became less popular. 
Recently, \citet{granot2021drop} revived classical patch-based approaches for image synthesis, and
was shown to significantly outperform \emph{single-image} GANs in both run-time and visual quality.

In light of the above-mentioned deficiencies of single-video GANs, and inspired by~\cite{granot2021drop}, 
we propose VGPNN (\emph{Video Generative Patch Nearest Neighbors}), a fast and practical method for video generation and manipulation from a single video. 
In order to handle the huge amounts of space-time patches in a single video sequence, we use and extend classical fast approximate nearest neighbor search methods~\cite{barnes2009patchmatch}.
We also employ robust optical-flow-like descriptors, which allow transferring highly different dynamics and motions from one video to another. Finally, by adding stochastic noise to the process, our approach can generate a large diversity of random different video outputs from a single input video in an unconditional manner.

Like single-video GANs, our approach enables the diverse and random generation of videos. However, in contrast to existing single-video GANs, VGPNN can generate \emph{high resolution} videos, while reducing runtime by many orders of magnitude, thus making diverse unconditional video generation from a single video realistically possible for the first time. 

\vspace*{0.05cm}
\noindent \underline{Our contributions are as follows:}
\begin{itemize} [topsep=0pt,itemsep=-1ex,partopsep=1ex,parsep=1ex,leftmargin=*]
\item We observe that space-time patch nearest-neighbor approaches, when posed as an unconditional generative model, outperform single-video GANs by a large margin, both in runtime and in quality.
\item We provide a new baseline for comparing single-video generative models. Our approach is practical to run and compare against (code will be released), and scalable also to high resolution videos (spatial or temporal).
\item We demonstrate a variety of video synthesis tasks, all within a single unified framework. These include: diverse unconditional video generation, video analogies, sketch-to-video, spatio-temporal video retargeting and conditional video inpainting.
\end{itemize}

\section{Related work}

Classical video generation methods, many of whom inspired by similar \emph{image} methods \cite{efros1999texture, efros2001image, wei2000fast}, include video texture synthesis \cite{kwatra2003graphcut,kwatra2005texture,kwatra2007texturing}, MRF-based controllable synthesis \cite{schodl2000video}, flow-guided synthesis \cite{bhat2004flow,kwatra2007texturing,jamrivska2015lazyfluids,okabe2009animating,sato2018editing,sato2018example} and more (see surveys by \cite{wei2009state,barnes2015patchtable}). While some used a generative model to model patch distribution, none considered unconditional generation of natural videos, beyond dynamic textures.

PatchMatch by \citet{barnes2009patchmatch} is a fast method for finding an approximate nearest-neighbor field (NNF). On natural images/videos, the algorithm converges very quickly to a good approximated solution. PatchMatch was also extended for k-nearest neighbors search \cite{barnes2010generalized}, faster search \cite{barnes2015patchtable} and being differentiably learnable \cite{duggal2019deeppruner}. The main editing tool shown in \cite{barnes2009patchmatch} is image summarization using bidirectional similarity (BDS). Introduced by \citet{simakov2008summarizing}, BDS ensures both visual ``completeness'' and ``coherence'' of the visual summary (coherence is obtained when the generated output contains only patches from the input, while completeness is obtained when all the patches in the input can be found in the generated output).
\cite{granot2021drop} proposed a normalized score to encourage completeness by globally re-weighting the similarity score between patches. Motivated by this similarity score, we utilize PatchMatch to efficiently incorporate global patch-specific information.

\section{Method} 
\label{sec:method}

\begin{figure}
    \includegraphics[width=\columnwidth]{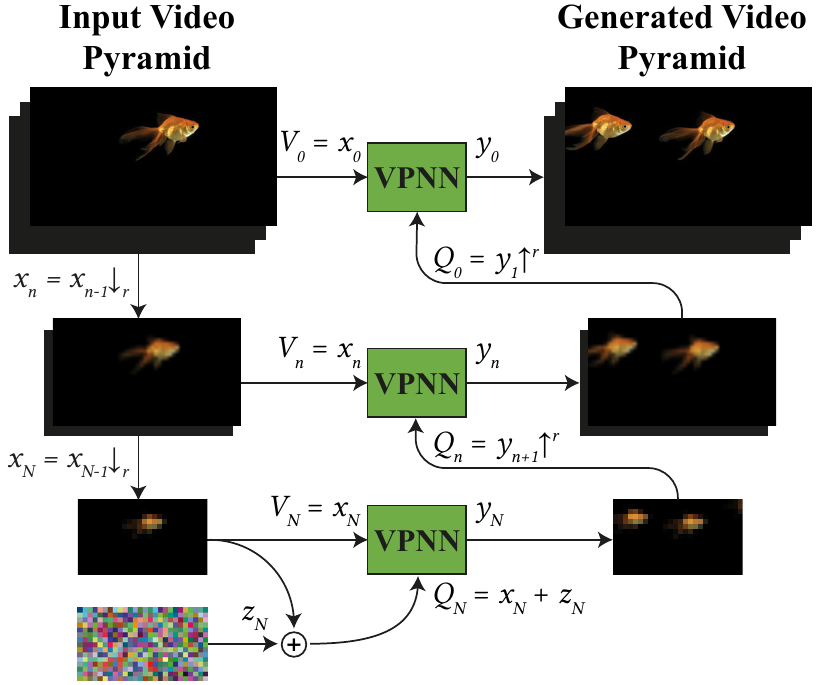}
    \caption{\textbf{VGPNN Architecture:} Given a single input video $x_0$, a spatio-temporal pyramid is constructed and an output video $y_0$ is generated coarse-to-fine. At each scale, VPNN module (see Fig.~\ref{fig:method_vpnn}) is applied to transfer an initial guess $Q_n$ to the output $y_n$ which shares the same space-time patch distribution as the input $x_n$. At the coarsest scale, noise is injected to induce spatial and temporal randomness.
    \vspace{-12pt}
    }
    \label{fig:method_pyramid}
\end{figure}

Our goal is to generate diverse video samples based on a single input video. We want our model to operate on natural input videos that can vary in their appearance and dynamics. In order to capture both spatial and temporal information of a single video, we start by building a spatio-temporal pyramid and operate coarse-to-fine to capture the internal statistics of the input video at multiple scales (Fig.~\ref{fig:method_pyramid}).
This multi-scale approach is extensively used in classical image synthesis methods as well as in modern GANs \citep[e.g.,][]{karras2017progressive,shaham2019singan,gur2020hierarchical}). At each scale we employ a Video-Patch-Nearest-Neighbor module or \textit{VPNN} (VGPNN is in fact a sequence of VPNN layers). The inputs to each layer depend on the application, where we first focus on our main application of diverse video generation (see Sec.~\ref{sec:applications} for the specific details of the other applications).

\paragraph{Multi-scale approach (Fig.\ref{fig:method_pyramid}):}
Given an input video $x$, we construct a spatio-temporal pyramid $\left\{x_0 \dots, x_N\right\}$, where $x_0=x$, and $x_n = x_{n-1} {\downarrow_{r}}$ is a bicubically downscaled version of $x_{n-1}$ by factor $r$ (${r=(r_H,r_W,r_T)}$, where $r_H=r_W$ are the spatial factors and $r_T$ is the temporal factor, which can be different). 

At the coarsest scale, the input to the first VPNN layer is an initial coarse guess of the output video. This is created by adding random Gaussian noise $z_N$ to $x_N$.
The noise $z_N$ promotes high diversity in the generated output samples from the single input. The global structure (e.g., a head is above the body) and global motion (e.g., humans walk forward), is prompted by $x_N$, where such structure and motion can be captured by \emph{small space-time} patches. The coarsest-scale output $y_N$ is generated by replacing each space-time patch of the initial coarse guess ($x_N + z_N$) with its nearest neighbor patch from the corresponding coarse input $x_N$. The resulting patches are then folded to a video, by choosing at each space-time position the median of all suggestions from neighboring patches.

At each subsequent scale, the input to the VPNN layer is the bicubically-upscaled output of the previous layer (${y_{n+1}\uparrow}^r$). The output $y_n$ is then generated by replacing each space-time patch with its nearest neighbor patch from the corresponding input $x_n$ (using the same patch-size as before, now capturing finer details).
This way, the output $y_n$ in each scale is similar in structure and in motion to the initial guess, but contains the same space-time patch statistics of the corresponding input $x_n$. Finally, the resulting patches are folded to a video. 

To further improve the quality and sharpness of the generated output at each pyramid scale ($y_n$), we iterate several times through the current scale VPNN layer, each time using the current output $y_n$ as input to the next iteration (similar to the EM-like approach employed in many patch-based works \citep[e.g.,][]{granot2021drop,simakov2008summarizing,barnes2009patchmatch,wexler2004space}).

\begin{figure}
    \includegraphics[width=\columnwidth]{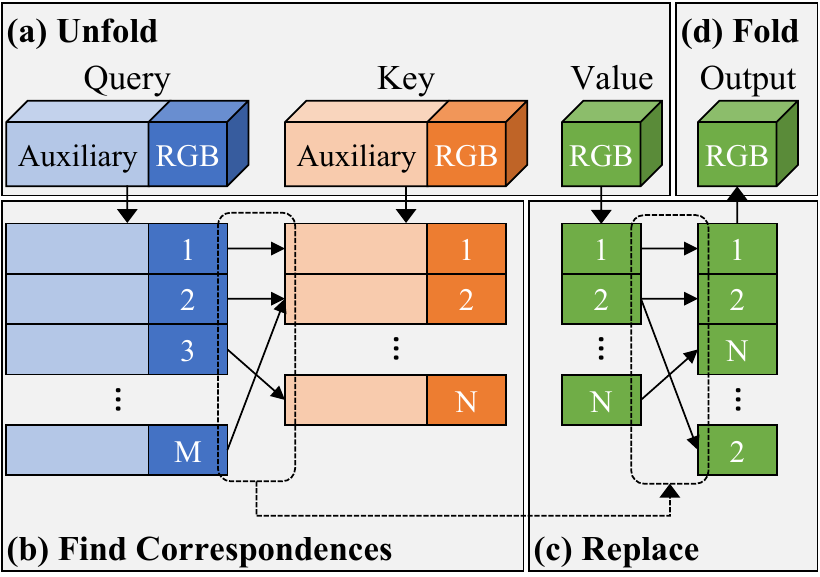}
  \caption{\textbf{VPNN module} 
   gets as input RGB videos query, key and value (QKV respectively). Q and K can be concatenated to additional auxiliary channels. It outputs an RGB video. (a) Inputs are unfolded to patches (each position now holds a concatenation of neighboring positions); (b) Each patch in Q finds its nearest neighbor patch in K. This is achieved by solving the NNF using PatchMatch \cite{barnes2009patchmatch}; (c) Each patch in Q is replaced with a patch from V, according to the correspondences found in stage (b); (d) Resulting patches are folded back to an RGB video output.
  }
  \label{fig:method_vpnn}
\end{figure}

\paragraph{QKV scheme (Fig.~\ref{fig:method_vpnn}):} Similar to \cite{granot2021drop}, we adopt a QKV scheme (query, key and value, respectively;  as is common in the attention mechanism~\cite{vaswani2017attention} nomenclature) for VPNN.
Instead of comparing two patches by only using their RGB values, in several cases it is necessary to compare patches in another search space. 
For example, using the aforementioned notations, we denote $V=x_n$ (the corresponding level from the pyramid of the original video) and $Q={y_{n+1}\uparrow}$ (the upscaled output of previous layer). Note that since $Q$ is an \emph{upscaled} version of previous output, its patches are blurry. Seeking their nearest neighbors in $V$ (whose patches are sharp) often results in improper matches. This is mitigated by setting $K=x_{n+1}{\uparrow^r}$ (in the first iteration of each scale of VGPNN), which has a similar degree of blur/degradation as $Q$. After finding its match in $K_j$, each patch $Q_i$ is then replaced with a patch $V_j$ (where $i,j$ are spatio-temporal positions. Also note that $K$ and $V$ are of the same shape). 

The QKV scheme is especially important in our video analogies application where it is used to include additional temporal information in the queries and the keys. We discuss it in detail in Sec.~\ref{sec:analogies}.

\paragraph{Completeness score:}
In the applications of video analogies, spatio-temporal video retargeting and conditional video inpainting we use the normalized similarity score \cite{granot2021drop} that encourages visual completeness.
The score between a query patch $Q_i$ and a key patch $K_j$ is defined as:
\begin{align}
    \label{eq:normalizedScore}
    S\left(Q_i, K_j\right) \coloneqq \frac{1}{\alpha + \min_\ell {D(Q_\ell, K_j)}} D(Q_i, K_j) \ \ ,
\end{align}
where $D=MSE$, and $\alpha$ controls the degree of completeness (smaller $\alpha$ encourages more completeness). $S$ is essentially a weighted version of $D$, whose weights depend \emph{globally} on $K$ and $Q$. 

\paragraph{Finding Correspondences:} 
\label{sec:patchmatch} 

We find the nearest neighbors between $Q$ and $K$ (Fig.~\ref{fig:method_vpnn}b) using PatchMatch~(\citet{barnes2009patchmatch}).
To cope with the completeness score, we apply PatchMatch twice. First we find a ``rareness" score for the keys - for each \emph{key} we find its closest \emph{query}. Then, for each \emph{query} we find its closest \emph{key} while factoring in the rareness of the keys as weights in the PatchMatch search.
Namely, we solve for:
\begin{align}
    \text{NNF}(\mathbf{p}) = \text{arg} \min_{\mathbf{v}} W(\mathbf{p+v}) \cdot D(Q(\mathbf{p}), K(\mathbf{p+v}))
\end{align}
Where $D$ is a distance function, $W$ are per-patch weights, $\mathbf{p}=(t, x, y)$ a position in $Q$ and $\mathbf{v}=(t',x',y')$ are possible NNF candidates (such as the NNF at the current position $\text{NNF}(t,x,y)$ or at a neighbor position $\text{NNF}(t,x-1,y)$ in the propagation step). 

This requires a slight modification of PatchMatch to support per-key weights. This additional support makes it possible to approximately solve Eq.~\ref{eq:normalizedScore} with two passes of PatchMatch. Even though this gives an approximation of Eq.~\ref{eq:normalizedScore}, we do not suffer loss in quality or lack of completeness, as apparent from our results.

The algorithm is implemented on GPU using PyTorch~\cite{NEURIPS2019_9015}, with time complexity $O(n \times d)$ and $O(n)$ additional memory (where $n$ is the video size and $d$ is the patch size; also see Fig.~\ref{fig:runtime_single}).

\paragraph{Temporal Diversity and Consistency:}
To enhance the temporal diversity of our samples we set the temporal dimension of the output to be slightly smaller than that of the input video. Thus, motions in different spatial positions in the generated output are taken from different temporal positions in the input video, increasing the overall temporal diversity (see for example the generated dancers in Fig.~\ref{fig:generation_examples} that are not synced).
We also found that the temporal consistency is best preserved in the generated output when the initial noise $z_N$ is randomized for each spatial position, but is the same (replicated) in the temporal dimension.


\section{Experimental Results}
\label{sec:experiments}

In this section we evaluate and compare the performance of our main application -- diverse video generation from a single input video. 
Figs.~\ref{fig:teaser} and~\ref{fig:generation_examples} illustrate 
diverse videos generated from a single input video, all sharing the same space-time patch distribution. The diversity is both spatially (e.g., number of dancers and their positions are different from the input video) and temporally (generated dancers are not synced).
Please refer to the supplementary material to view the full resolution videos and many more examples.

\paragraph{Comparison to other methods for video generation from a single video:}
We compare our method to recently published methods for diverse video generation from single video: \hpvaegan~\cite{gur2020hierarchical} and SinGAN-GIF \cite{arora2021singan}.
We show that our results are both qualitatively and quantitatively superior while reducing the runtime by a factor of $3 \times 10^4$ (from 8 days training on one video to 18 seconds for new generated video).
Since SinGAN-GIF did not make their code available, and the training time of \hpvaegan for a single video (13 frames of size $144{\times}256$) is roughly $8$ days, we are only able to compare to the videos published by these methods.
``\hpvaegan dataset'' comprises of 10 input videos with 13 frames each, and of spatial resolution of $144{\times}256$ pixels. ``SinGAN-GIF dataset'' has 5 input videos with maximal resolution of $168{\times}298$ pixels and 8-16 frames.

\begin{figure}
  \includegraphics[width=\columnwidth]{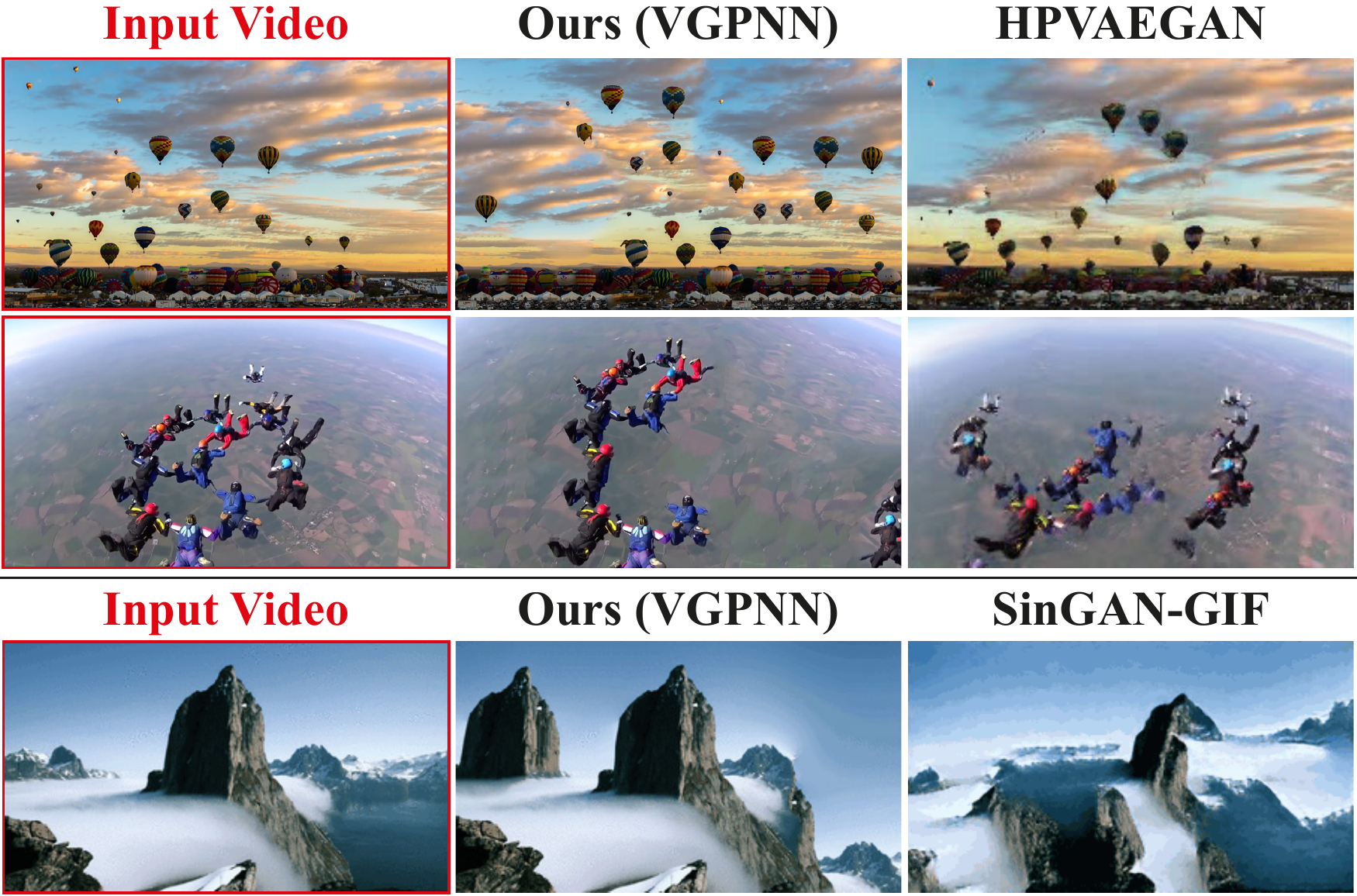}
  \caption{
  \textbf{Comparing Visual Quality} between our generated frames and those of \hpvaegan \cite{gur2020hierarchical} and SinGAN-GIF \cite{arora2021singan} (please \textbf{zoom in} on the frames). Note that our generated frames are sharper and also exhibit more coherent and plausible arrangements of the scene. For details see Section~\ref{sec:experiments}. Please find the full videos and more comparisons in the supplementary material.}
  \vspace{-12pt}
  \label{fig:generation_ours_vs_hpvaegan_singangif}
\end{figure}

\paragraph{Qualitative comparison:}
In Fig.~\ref{fig:generation_ours_vs_hpvaegan_singangif} we show a side-by-side comparison of representative generated frames of our method to frames generated by \hpvaegan \cite{gur2020hierarchical} and SinGAN-GIF \cite{arora2021singan}. Note that while \cite{gur2020hierarchical,arora2021singan} are limited to generated outputs of small resolution ($144{\times}256$), we can generate outputs in the same resolution of the input video (full-HD $1280{\times}1920$, shown in the figure). The full videos (as well as a comparison to our generated outputs of similar low resolution) can be viewed in the supplementary material. As can be seen, our generated samples (in low and high resolution) are more spatially and temporally coherent, as well as having higher visual quality. It is evident that generating videos using the space-time patches of the original input video, rather than regressing output RGB values, gives rise to high quality outputs.

\begin{figure}
    \centering
    \begin{tabular}{c}
        \includegraphics[width=\columnwidth]{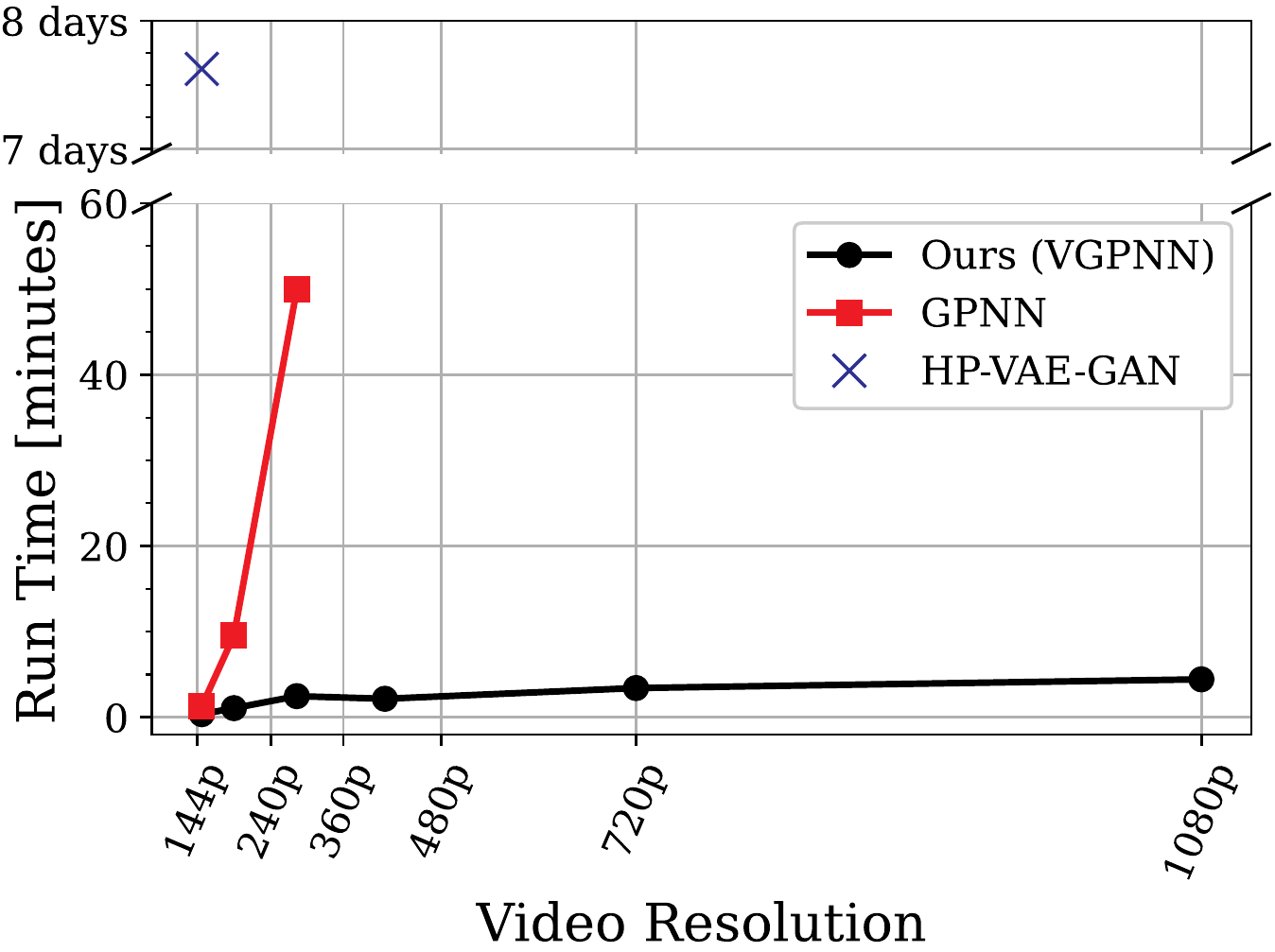}
    \end{tabular}
    \caption{\textbf{Comparing Generation Runtime} between our approach (VGPNN), a na\"ive extension of GPNN~\cite{granot2021drop} from 2D to 3D and \hpvaegan \cite{gur2020hierarchical}. We compared the generation time of 13-frames videos with different spatial resolutions (X-axis). All videos have 16:9 aspect ratio (e.g., 144p is 144x256 and 1080p is 1080x1920 -- full-HD). See Section~\ref{sec:experiments} for details.
    }
    \label{fig:runtime_single}
\end{figure}

\paragraph{Quantitative comparison:}
In Table~\ref{table:comparison} we report the Single-Video-FID (SVFID)~\cite{arora2021singan} of our generated samples, compared to those generated by \hpvaegan~\cite{gur2020hierarchical} and SinGAN-GIF~\cite{arora2021singan}\footnote{All quantitative comparisons were done on generated samples of the same resolution and video length as that of the other method.}. SVFID was proposed by~\cite{gur2020hierarchical} to measure the patch statistics similarity between the input video and a generated video. It computes the Fréchet distance between the statistics of the input video and the generated video using pre-computed C3D~\cite{tran2015learning} features (Lower \mbox{SVFID} is better). 
As can be seen in Table~\ref{table:comparison}, our generated samples bear more substantial similarity to the input videos (indicated by lower SVFID).
\cite{shaham2019singan} proposed a diversity index to make sure that generated outputs are indeed different (and not simply ``copying'' the input). We adapt the index for videos. The index is zero if all generated outputs are the same, and higher otherwise. While our and \hpvaegan generated samples have similar index ($0.45/0.41$ respectively), those of SinGAN-GIF have higher index ($0.86$ vs. our $0.6$). Such high diversity is not an advantage, when paired with SVFID about twice worse than ours. It stems from low quality appearance with out-of-distribution patches.
All inputs and generated videos can be found in the supplementary material.

\paragraph{User study:}
We conducted a user study evaluation using Amazon Mechanical Turk (AMT). For each dataset, 100 subjects were shown multiple pairs of videos, each consisting of a video generated by our method, and a video generated by the other method (both were generated from the same input video). The subjects were asked to judge which sample is better in terms of sharpness, natural look and coherence. In Table~\ref{table:comparison} we report the percentage of users who favored our method over the other. Compared to videos generated from \hpvaegan dataset, there is a clear preference in favor of our patch-based method. The results on the SinGAN-GIF dataset are not that clear-cut, this might be due to the somewhat restricted nature of the videos in that particular dataset (as mentioned above, it was not possible to check SinGAN-GIF on other samples, since the authors did not publish their code, nor stated the amount of time it took to generate their samples).

\paragraph{Reducing running times:}
In Fig.~\ref{fig:runtime_single} we show a comparison of the runtime taken to generate random video samples using our method, compared to a na\"ive extension of GPNN \cite{granot2021drop} (from 2D to 3D patches) and compared to the training time of \hpvaegan \cite{gur2020hierarchical}. As discussed in Sec.~\ref{sec:method}, the use of efficient PatchMatch algorithm for nearest neighbors search, as opposed to the exhaustive search done in GPNN, dramatically reduces both run time and memory footprint used for video generation, making it possible to generate high-resolution videos (including Full-HD 1080p). All experiments were conducted on Quadro RTX 8000 GPU.

\begin{table}[t]
\begin{adjustbox}{max width=\columnwidth}
\begin{tabular}{lccc}
\toprule

\textbf{Method} 
&
\textbf{SVFID} 
&
\textbf{Head-on comparison}
&
\textbf{Runtime}   
\\
&
\cite{gur2020hierarchical} $\downarrow$
&
(User study) \textbf{[$\%$]}$\uparrow$
&
$\downarrow$
\\

\midrule

 \hpvaegan \cite{gur2020hierarchical}     & 0.0081  & \multirow{2}{*}{\textbf{67.84}$\pm$1.77} &    7.625 days \\ 
 \textbf{VGPNN} (Ours)  & {\textbf{0.0072}} &    &  \textbf{18} secs \\
 
\midrule
 
 SinGAN-GIF \cite{arora2021singan}  & 0.0119   &  \multirow{2}{*}{\textbf{50.57}$\pm$ 3.27} &  Unpublished \\ 
 \textbf{VGPNN} (Ours) & {\textbf{0.0058}} &  &  \textbf{10} secs \\

\bottomrule

\end{tabular}
\end{adjustbox}
\caption{\textbf{Quantitative Evaluation:} A comparison of our generated video samples to that of \hpvaegan~\cite{gur2020hierarchical} and SinGAN-GIF~\cite{arora2021singan}, conducted on input videos provided in their papers. Our diverse samples have more resemblance to the input videos (indicated by lower SVFID). In a user study, users scored in favor of our method (see Section~\ref{sec:experiments} for details).
}
\label{table:comparison}

\end{table}

\section{Applications}
\label{sec:applications}

Other than unconditional diverse generation, we demonstrate the utility of the proposed unified framework on several other video manipulation applications.

\begin{figure*}
  \includegraphics[width=\textwidth]{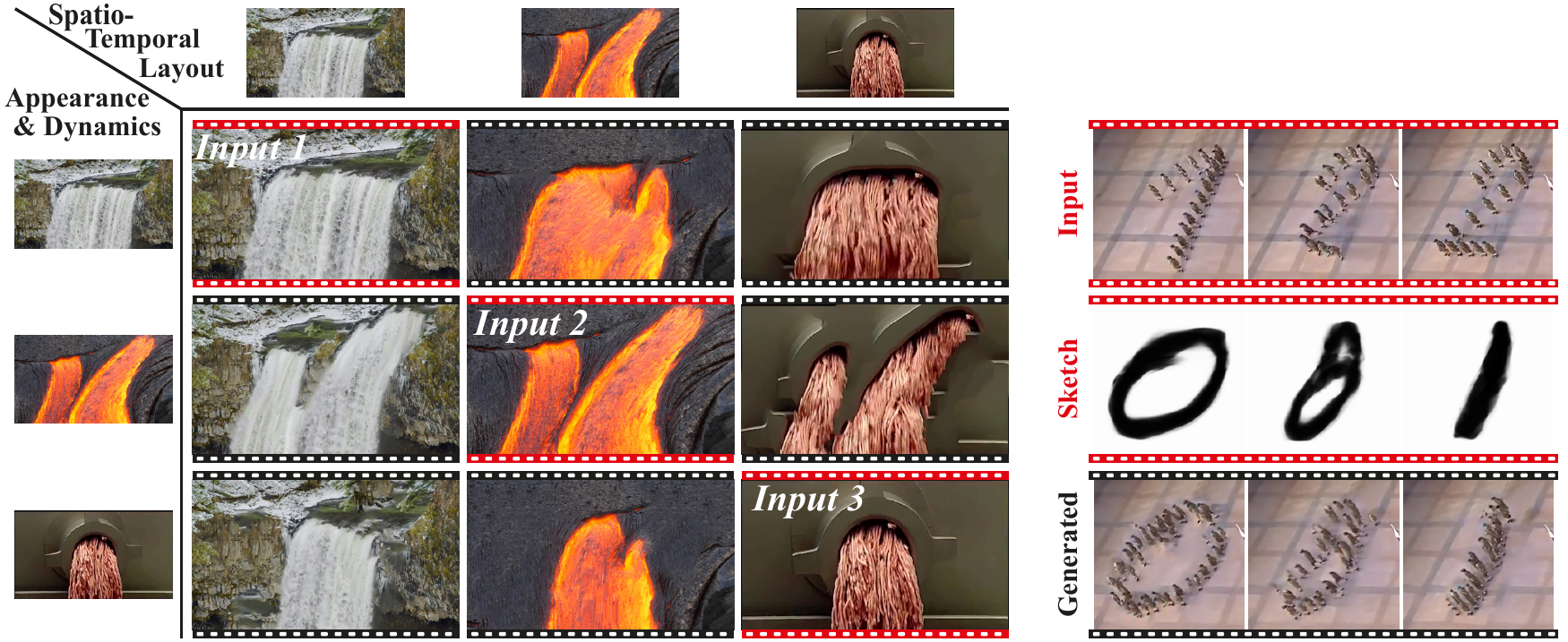}
  \caption{
  \textbf{Video Analogies:} \textit{Left:} an example of video analogies between all pairs of three input videos (red). Each generated video (black) takes the spatio-temporal layout from the input video in its row, and the appearance and dynamics of the input video from its column. \textit{Right:} an example of sketch-to-video -- the generated video (bottom) takes its spatio-temporal layout from the sketch video of morphed MNIST digits (middle) and its appearance and dynamics from the input video of parading soldiers (top). \textit{Please find full videos and additional examples in the supplementary material}.
  }
  \label{fig:analogies}
\end{figure*}

\subsection{Video analogies}
\label{sec:analogies}
Inspired by \textit{Image Analogies} \cite{hertzmann2001image,liao2017visual,benaim2021structural} we propose \textit{Video Analogies} where we generate a new video whose spatio-temporal layout is taken from a content video $C$, and overall appearance and dynamics are taken from a style video $S$ (see Fig.~\ref{fig:analogies}).

Our goal is to find a mapping of dynamic elements (patches) between the two videos, which can be very different in their appearance (RGB space). This is achieved by using the magnitude of the optical flow (extracted via RAFT~\cite{teed2020raft}), quantized into few bins (using k-means)\footnote{Each cluster has an integer cluster index. We divide each index by the total number of clusters/bins to be in $[0,1]$}. We term this the \textit{dynamic structure} of the video. By concatenating the dynamic structure to the RGB values of the video (along the channels axis), each patch can now be compared using its RGB values and its dynamic values. This provides a good mapping between the dynamic elements of the two input videos.

Given the dynamic structures of both input videos $\text{dyn}(C)$, $\text{dyn}(S)$, we compute their spatio-temporal pyramids and that of the style video $S$. The output video is generated by setting $Q,K,V$ at each scale as follows:

\begin{flushleft}
$
\begin{array}{@{}l|lll}
     \text{Scale} & Q & K & V \\
     \midrule
     \text{N (coarsest)}& \text{dyn}(C)_N & \text{dyn}(S)_N & S_N \\
     \text{n (any other)} & \text{dyn}(C)_n \Vert Q_{n+1}\uparrow & \text{dyn}(S)_n  \Vert S_n & S_n 
\end{array}
$
\end{flushleft}
where $\Vert$ denotes concatenation along the channels axis, and $n$ denote the current scale in the pyramid. Note that in the coarsest scale, the two videos are only compared by their dynamic structure. In finer scales, the dynamic structure of $C$ (the content video) is used to ``guide'' the output to the desired spatio-temporal layout.

In Fig.~\ref{fig:teaser} we show a snapshot of the analogies between a waterfall and a lava stream, and in Fig.~\ref{fig:analogies} we show snapshots of the analogies of all possible pairs between three videos (the lava stream, a waterfall and a meat grinder). \textit{The full videos are in the supplementary material}. 

We can use the above mentioned mechanism for ``sketch-to-video'' transfer, where the dynamic structure is given by a sketch video instead of an actual video. See Fig.~\ref{fig:analogies} for a few snapshots of transferring the motion of morphed MNIST~\cite{lecun1998mnist} digits to a video of marching soldiers, and \textit{please see the full videos and many more results in the supplementary material}.

Related to us are works for video style transfer from a style image \citep[e.g.][]{benard2013stylizing,ruder2016artistic,jamrivska2019stylizing,chen2017coherent}, general video-to-video translation trained on large datasets or conditioned on human poses or keypoint detection \citep[e.g.][]{wang2018video, wang2019few, mallya2020world, wang2020imaginator} or human motion transfer methods \citep[e.g.][]{aberman2020unpaired, chan2019everybody,yang2020transmomo,siarohin2019animating,siarohin2019first,ren2020human,lee2019metapix} that involve some kind of human model knowledge. Flow-based appearance transfer of fluids has been studied by \cite{bhat2004flow,kwatra2007texturing,jamrivska2015lazyfluids,okabe2009animating,sato2018editing,sato2018example}. Most similar to us is \cite{jamrivska2015lazyfluids} that uses a patch nearest neighbor approach to transfer the appearance of a fluid exemplar (a still image) into a video given a human annotated flow+alpha mask. Our method differs in how we model the flow guidance and in the mapping we have between two flows of two videos (instead of a still image exemplar).

Also similar to us is Recycle-GAN by \citet{bansal2018recycle} that pose  unsupervised video-to-video translation as a domain transfer problem (each video is a domain). They train convolutional encoders to map between the two videos using adversarial loss with cyclic constraints.
We provide a comparison to \cite{bansal2018recycle} in the supplementary material. As can be seen for the sketch to videos examples, RecycleGAN's outputs fail to capture the finer motions of the soldiers, and in the fluid-like examples it quickly converges to the style video. On top of that, the overall visual quality is diminished due to the model being parametric.

\subsection{Spatial Retargeting} 
The goal of video retargeting is to change the dimensions of a video without distorting its visual contents (e.g., fit a portrait video to a wide screen display). It can be performed in a very similar manner to our video generation described in Sec.\ref{sec:method}. Given a target shape, we first resize (bicubically) the input video to the target shape, then compute two pyramids (for the input and resized videos) with the same depth and downscale factor. The initial guess at the coarsest scale $Q_N$ would be the coarsest scale of the resized pyramid (without any additional noise). We then compute the rest of the output video in the same manner as in Sec.\ref{sec:method}. Note that at each scale, $V_n$ are unchanged, hence no distortion is introduced to the patches reconstructing the retargeted video.

As can be seen in Fig.~\ref{fig:teaser} and in the supplementary material, the results preserve the original size and aspect ratio of objects from the input videos while keeping the overall appearance coherent even though the aspect ratio is significantly altered. The dynamics and motions in the videos are also preserved. For instance, the balloons are not ``squashed" but rather packed more compactly in the sky and more members were added to the choir instead of stretching them. Nevertheless, the motion of the balloons or the sway of the choir members are preserved. 
Other classical works for video retargeting, such as \cite{simakov2008summarizing, wolf2007videoretarget, rubinstein2008improved,krahenbuhl2009disney} did not make their implementation available, therefore we were unable to provide a comparison.

\subsection{Temporal Retargeting} 

Similar to spatial retargeting, one can generate a realistic video with a different \emph{temporal} length. One possible use is generating a shorter summary of the video.
While most deep video summarization techniques are achieved by selecting a subset of frames (see survey~\cite{apostolidis2021video}), classical methods have demonstrated summaries that consist of \emph{novel frames} in which dynamics that are originally sequential can be parallelized or vice versa \cite{rav2006making, simakov2008summarizing}. By applying the retargeting approach to the temporal dimension, VGPNN generates summaries with novel frames. 
The \emph{temporal retargeting section in the supplementary material} shows several examples.
For example, in the dog training summarized video, the trainer and dog turn around simultaneously as opposed to sequentially in the original video.
Moreover, we can, in a similar manner, extend the temporal duration of a video creating longer dynamics while preserving the speed of the individual actions.
In the ballet dancer video for example, the choreography is longer, but the pace of the dance motions remains the same. 

\subsection{Video conditional inpainting}
\begin{figure}
  \includegraphics[width=\columnwidth]{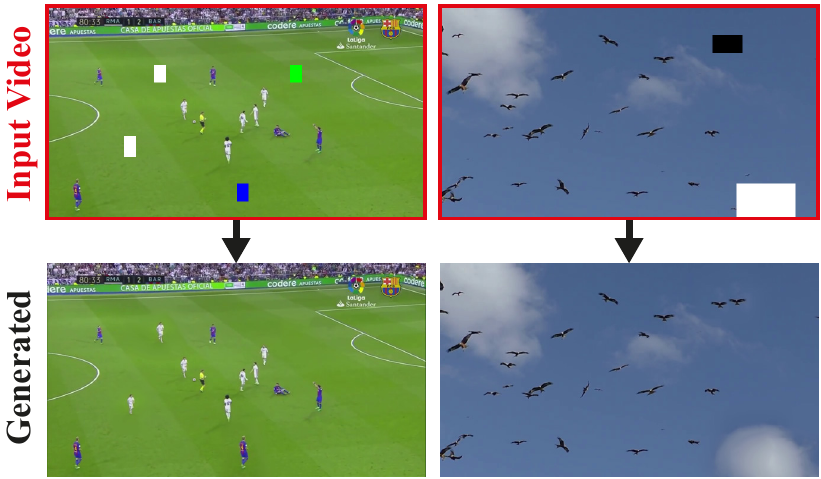}
  \caption{\textbf{Conditional Video Inpainting.}
  Colored occluded masks in the input video (red) are completed, conditioned on the color cue. The completion persists through the video dynamics. Please see full videos in the supplementary material.}
  \label{fig:conditional_inpainting}
\end{figure}

In this task we are given an input video with some occluded space-time volume, where the missing parts should be completed based on crude color cues placed by the user in the occluded space (similar to conditional image inpainting \cite{granot2021drop}). Here we set the number of levels in the pyramid such that the occluded part in the coarsest scale is roughly in the size of a single patch. The masked part is then coherently reconstructed using other space-time patches of similar colors to that of the cue. In finer scales, details and dynamic elements are added correctly. Fig.~\ref{fig:conditional_inpainting} shows how different cues are completed with different elements from the non-occluded parts. See for instance, how a blue cue will be replaced by a player from Barcelona while a white cue by a player from Real Madrid. See more examples in the supplementary material.

\section{Limitations}
Generation of local patches, as in VGPNN but also as in single video GANs, lacks global geometric consistency. This is apparent when scenes with significant depth variations are introduced with large camera motion. While each frame is plausible, different patches are not being transformed consistently, resulting in non-rigid deformations to entities that are realistically rigid. See the generated videos of mountains in the supplementary. 

\section{Conclusion}
We demonstrated that random diverse video generation from a single video can be efficiently done by simple patch-based methods. We also demonstrated how small modifications to our framework give rise to other tasks such as video analogies and spatio-temporal retargeting. We showed that our non-parametric approach outperforms existing single-video GANs in the visual quality of the generated outputs, while being orders of magnitude faster. 
The low run time required for generating videos using our approach makes it a good baseline for future works in the field.

\section*{Acknowledgements}
This project received funding from the European Research Council (ERC) under the European Union’s Horizon 2020 research and innovation programme (grant agreement No 788535), from the D. Dan and Betty Kahn Foundation, and from the Israel Science Foundation (grant 2303/20). Dr. Bagon is a Robin Chemers Neustein AI Fellow.


{\small
\bibliographystyle{abbrvnat}
\bibliography{vgpnn}
}

\newpage

\appendix

\section{Implementation Details}

All RGB values in the videos are scaled to $[-1,1]$. All runs were conducted on Quadro RTX 8000 GPU.

\paragraph{Creating the Spatio-Temporal Pyramid}
Given a downscaling factor $r$ (where $r=(r_H, r_W, r_T)$) and a minimal size $m_S,m_T$, we keep downscaling the input video in all dimensions until it ``hits" the minimal size in the spatial or temporal dimensions. Assume we hit the minimal spatial dimension first, we keep on downscaling the temporal dimension until reaching its minimal size, while keeping the spatial dimensions fixed on its minimal size (and the opposite goes if we first hit the temporal dimension, keeping on downscaling the spatial dimensions while keeping the temporal fixed). The minimal size of the spatial dimensions, $m_S$, is the minimum between both height and width (namely, no spatial dimension will be smaller than $m_S$).

We use cubic downscaling interpolation for both temporal and sptial dimensions. We tried to use nearest interpolation on the temporal dimension, because it might make more sense sometimes, but found that in most applications it performed the same or worse.

\paragraph{Diverse Generation}
We used Gaussian noise with standard deviation of $2-5$. Downscaling factor is $0.82$ for the spatial dimensions (height and width) and $0.87$ for the temporal dimension. The minimal size of the pyramid is set to $3$ frames with minimal spatial dimension of $15$ pixels. We use patch size $(3\times7\times7)$, where~3 is in the temporal dimension. We use $5$ EM-like iterations in each scale of the pyramid. When the number of voxels ($T\times H\times W$) is larger than $3,000,000$ we change the number of EM-like iterations to $1$, and the patch-size to $(3\times5\times5)$. This change reduce runtime without hurting the quality of the results. 

\paragraph{Video Analogies}
For all examples we use patch size $(3\times5\times5)$ and $\alpha=1$ (for completeness score).
For all-pairs examples we use downscaling factor is $0.9$ for all dimensions. The minimal size of the pyramid is set to $3$ frames with minimal spatial dimension of $20$ pixels. $1$ EM-like iterations per scale. For sketch-to-video examples we use downscaling factor of $0.78$ for all dimensions and minimal size is $5$ frames and minimal spatial dimension of $35$. $3$ EM-like iterations per scale (and $1$ for the last two scales, to save runtime). Runtime per result is about 1 minute.

\paragraph{Layout-Appearance Tradeoff in Video Analogies}
Since we are trying to create a new video whose spatio-temporal layout (modeled with the dynamic structure) taken from one video and its appearance from another, there's an inherent tradeoff of which of the two we want to be better preserved in the result.
The dynamic structure is ``enforced'' by using the auxiliary channels in $Q$ and $K$. Removing these channels would generate a video that is much more similar in its appearance to $S$ but bears less resemblance to the spatio-temporal layout of $C$.  We can control this tradeoff by setting an upper limit in the pyramid from which we stop using the auxiliary channels. In our results it was best to set the maximal scale at half the pyramid height.

\section{PatchMatch Implementation Details}
\begin{figure}
    \centering
    \begin{tabular}{c}
        \includegraphics[width=\columnwidth]{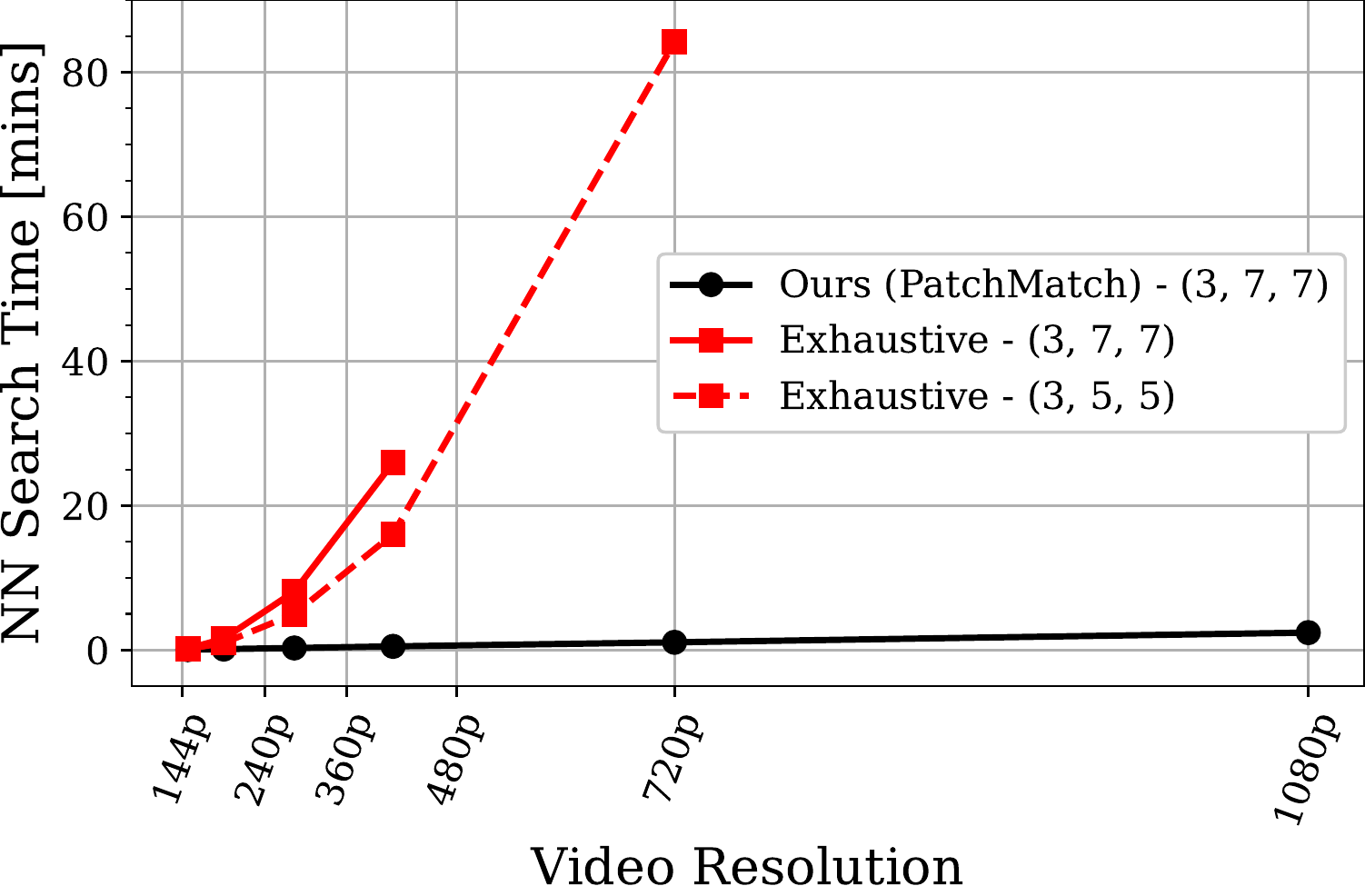} 
    \end{tabular}
    \caption{\textbf{Nearest Neighbor Search Comparison} using PatchMatch (in our method) vs. exhaustive search (used by GPNN \cite{granot2021drop}). GPNN exceeds GPU memory at medium resolution (480p) with the original patch size $(3,7,7)$. The dashed line with smaller patch size $(3,5,5)$ is intended to show the quadratic trend with more data points.
    }
    \label{fig:search_time}
\end{figure}

In all applications we use $L2$ as the distance function between. In Fig.~\ref{fig:search_time} we show another more detailed comparison between our PatchMatch implementation and the exhaustive nearest-neighbor search used by GPNN. Our implementation has time complexity of $O(n \times d)$ and $O(n)$ additional memory {(where $n$ is the video size and $d$ is the patch size)}, compared to GPNN, with time complexity of $O(n^2 \times d)$ and memory footprint $O(n \times d)$. This is easily seen in the figure.
We use the same propagation and random search steps as in the original PatchMatch paper \cite{barnes2009patchmatch}, using the ``jump flood'' scheme \cite{rong2006jump}. In each PatchMatch iteration we look at 4 neighbors at distance $step$ (with additional small noise for the exact position of the neighbor, and without) and a random search. However, we only use 15 PatchMatch iterations per VPNN usage, this is done by searching for $step=8,4,1$ 5 times.

\begin{figure}
    \centering
    \begin{tabular}{c}
        \includegraphics[width=\columnwidth]{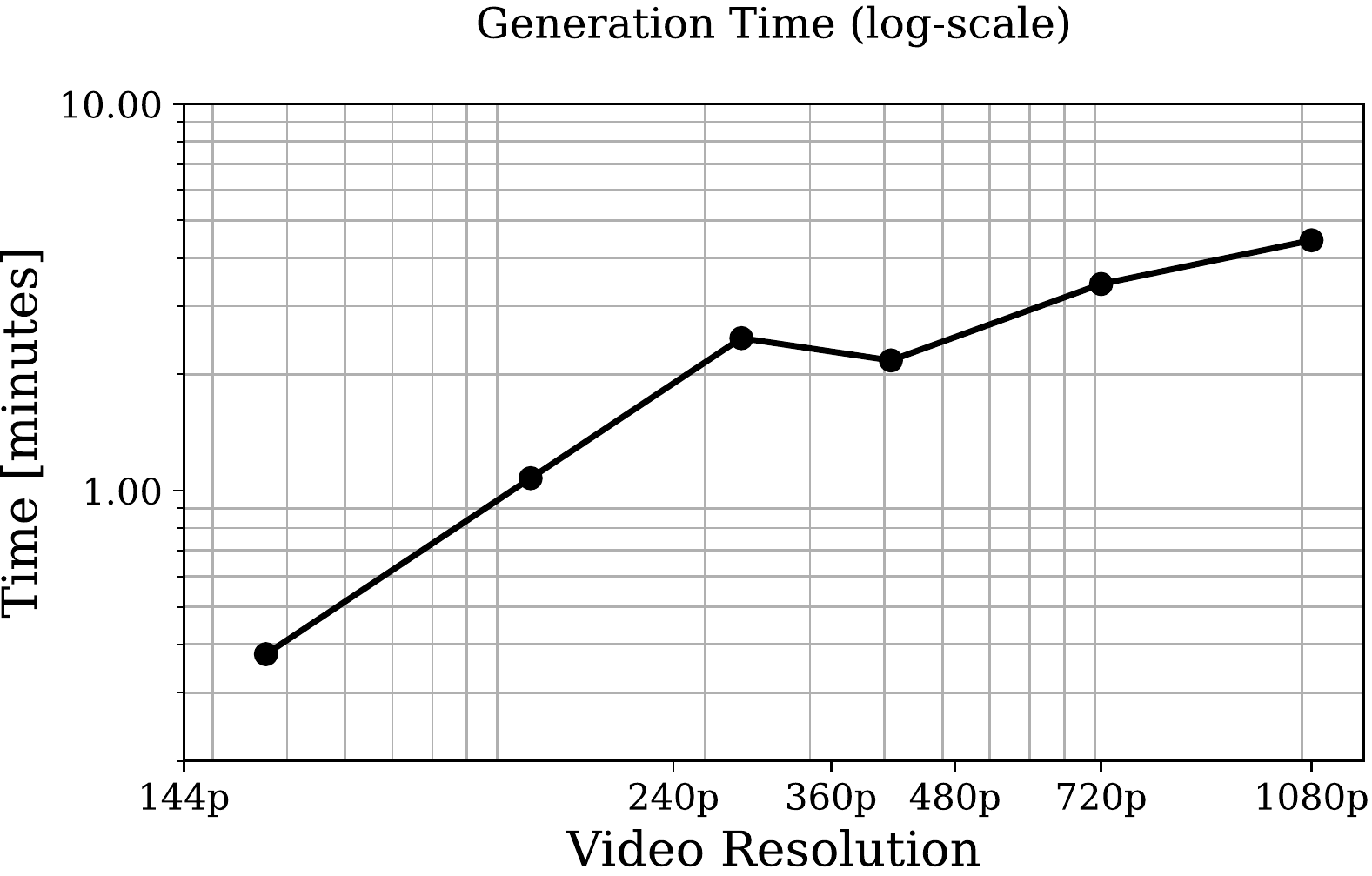} 
    \end{tabular}
    \caption{
    \textbf{Generation Time} Note y-axis is in log-scale. This is the same graph as in Fig.~\ref{fig:runtime_single}, shown here in log-scale so it would be easier to get a sense of the actual runtime needed to generate a random 13-frames video in the relevant resolution (x-axis).
    }
    \label{fig:runtime_ours}
\end{figure}

\section {Comparison details}
For each input video in \hpvaegan and SinGAN-GIF datasets we generated the same number of random sample as publicly available (10 generations for each video in \hpvaegan dataset, and 6 generations for each video in SinGAN-GIF dataset), and compared their SVFID and diversity. 

\paragraph{Video Diversity Index}
The video adaptation of the \textit{diversity} index (originally proposed for images by~\cite{shaham2019singan}) is: given an input video, the standard deviation of each video position (3D RGB element in the video, converted to grayscale) is calculated across all generated samples, and then averaged across all pixels. This is then divided by the standard deviation of the voxels in the input video.

\end{document}